\documentclass[runningheads]{llncs}
\usepackage[T1]{fontenc}
\usepackage{svg}
\usepackage{subcaption}
\usepackage{soul}
\usepackage{graphicx,verbatim}
\usepackage{booktabs}
\usepackage{multirow}
\usepackage{algorithmic}
\usepackage{array}
\usepackage{textcomp}
\usepackage{stfloats}
\usepackage{url}
\usepackage{verbatim}
\usepackage{graphicx}
\usepackage[english]{babel}
\usepackage[utf8]{inputenc}
\usepackage{algorithm}
\usepackage{diagbox}
\usepackage{siunitx}
\usepackage{lipsum}

\usepackage{amsmath,amssymb,amsfonts}

\usepackage{hyperref}

\begin{document}
\title{I$^2$-Med: Interpretable Medical Inference through Visual-Guided Dynamic Logits Calibration}
\titlerunning{I$^2$-Med}
%
\author{
Md Sajid\inst{1} \and
Shreeyut Maheshwari\inst{1} \and
Akshat Mishra\inst{1} \and
Ritik Mishra\inst{1} \and
Mohammad Tanveer\inst{1}
}

\authorrunning{M. Sajid et al.}

\institute{
Indian Institute of Technology Indore, Indore 453552, Madhya Pradesh, India\\
\email{\{phd2101241003,~me230003071\}@iiti.ac.in,~akshatm430@gmail.com, ~\{phd2301241003,~mtanveer\}@iiti.ac.in}
}


  
\maketitle              
\begin{abstract}
Medical vision language models have demonstrated strong potential for medical image understanding and reasoning. However, their interpretability and reliability remain limited because medical vision language tasks require precise visual grounding and clinically coherent explanations. This is particularly challenging due to complex imaging patterns and scarce expert annotations. These challenges reduce the effectiveness of conventional reasoning methods and lead to visual hallucination, defined as the generation of text that is not grounded in the given visual input, which further limits their applicability in clinical practice. To mitigate these challenges, we propose an interpretable medical inference (I$^2$-Med) framework that enhances visual faithfulness during inference without additional training to improve reasoning and explanation faithfulness in medical vision language models. During text generation, I$^2$-Med uses a visual guidance module to check whether each candidate word is supported by the medical image and consistent with the current reasoning. Based on this guidance, the model’s output scores are adjusted to favor visually grounded words, without changing the model parameters. Extensive experiments on the OmniMedVQA benchmark, covering eight medical imaging modalities and five clinical question types, demonstrate that our proposed I$^2$-Med framework generates more clinically meaningful and visually faithful reasoning chains and reduces visual hallucination under standard medical VQA evaluation metrics. The source code is publicly available at~
\url{https://github.com/mtanveer1/l2-med}.

\end{abstract}

\section{Introduction}
The recent advancement of Large Multimodal Models (LMMs) \cite{alayrac2022flamingo} has shown a paradigm shift in medical artificial intelligence, promising transformative applications in automated diagnosis, report generation, and clinical decision support \cite{radford2021learning,li2023llava}. In comparison to traditional deep learning models restricted to specific modalities or tasks, medical Vision Language Models (VLMs) \cite{pan2025medvlm} possess the capacity to interpret complex visual data and articulate findings through natural language. However, their deployment in clinical practice is hindered by two fundamental limitations: (i) the lack of transparent reasoning processes \cite{chen2025mitigating} and (ii) visual hallucination, the generation of plausible but image-ungrounded medical statements \cite{shao2024deepseekmathpushinglimitsmathematical}. In high-stakes medical settings, where interpretability and factual accuracy are paramount, such failures can lead to dangerous misdiagnoses and an erosion of clinical trust.

Current approaches to adapt VLMs to the medical domain primarily rely on Supervised Fine-tuning (SFT) on medical image-text pairs \cite{moor2023med}. While SFT enables models to learn domain-specific terminology, it often fails to instill robust reasoning capabilities. Models trained via SFT tend to overfit to the statistical correlations of the training data rather than learning to verify visual evidence, leading to superficial fluency without genuine understanding \cite{chu2025sft}. To address this, recent works have explored Reinforcement Learning (RL) strategies, such as Proximal Policy Optimization (PPO) and Group Relative Policy Optimization (GRPO), to align models with human preference and encourage Chain-of-Thought (CoT) reasoning \cite{shao2024deepseekmathpushinglimitsmathematical,lai2026med}. While RL effectively incentivizes structured reasoning (e.g., generating step-by-step explanations), it operates primarily on the textual reasoning level. Crucially, these methods optimize reasoning structure but do not explicitly enforce token-level alignment between generated explanations and the underlying medical image during inference. Consequently, even models capable of complex reasoning may still hallucinate details that do not exist in the medical data (e.g., MRI and PET scans) to maintain the logical flow of their self-generated text \cite{chen2025mitigating}.

    \begin{figure}
        \centering
        \includegraphics[width=1.0\linewidth]{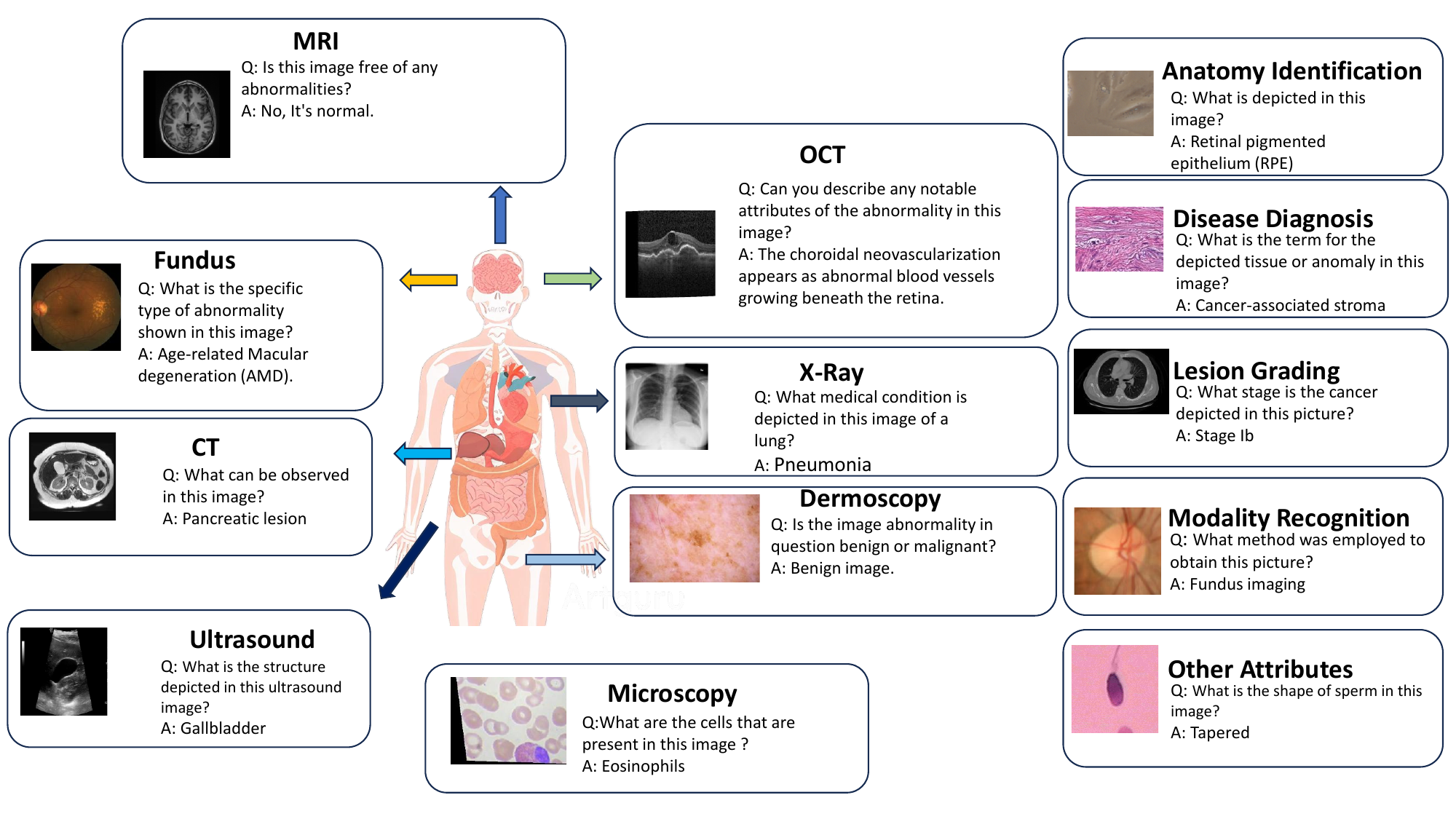}
        \caption{Overview of the Evaluation Framework: Eight Medical Imaging Modalities and Five Medical Vision Question Answering Tasks.}
        \label{fig:f1}
    \end{figure}

To bridge this gap between logical reasoning and visual faithfulness, we propose I$^2$-Med, a novel unified framework for Interpretable Inference in medical VLMs. I$^2$-Med addresses the hallucination problem by two key components: (i) a lightweight Reinforcement Learning strategy to induce structured reasoning, and (ii) a training-free visual guidance mechanism based on Dynamic Logits Calibration (DLC) for faithfulness enforcement. Unlike standard decoding methods that rely solely on the language model's probability distribution, our approach introduces a visual sanity check at the decoding stage. By leveraging BiomedCLIP \cite{zhang2023biomedclip}, a domain-specific contrastive vision language model, as an external visual alignment assessor, I$^2$-Med dynamically calibrates candidate token logits based on image-text consistency scores. This ensures that the generated text remains anchored to the visual evidence in real-time, penalizing tokens that drift from the image context without requiring computationally expensive retraining of the large backbone model. Our approach represents a significant step toward trustworthy medical AI by treating interpretability not as a post-hoc explanation, but as an intrinsic constraint of the generation process. The main contributions of this work are summarized as follows:
\begin{enumerate}

    \item We introduce I$^2$-Med, a unified framework that integrates GRPO-based structured reasoning with inference-time visual-logit calibration to jointly enhance reasoning quality and visual faithfulness.
    \item We propose MedCalib-IF, a training-free inference-time calibration mechanism that evaluates token-level visual grounding using BiomedCLIP and dynamically suppresses hallucinations through logits modulation.
    \item We conduct extensive experiments on the OmniMedVQA benchmark, covering eight imaging modalities and various clinical tasks (as shown in Figure \ref{fig:f1}). Our results demonstrate that I$^2$-Med significantly reduces visual hallucination and improves the clinical validity of generated explanations compared to state-of-the-art baselines.

\end{enumerate}


\section{The proposed Methodology}
\begin{figure}
    \centering
    \includegraphics[width=1\linewidth]{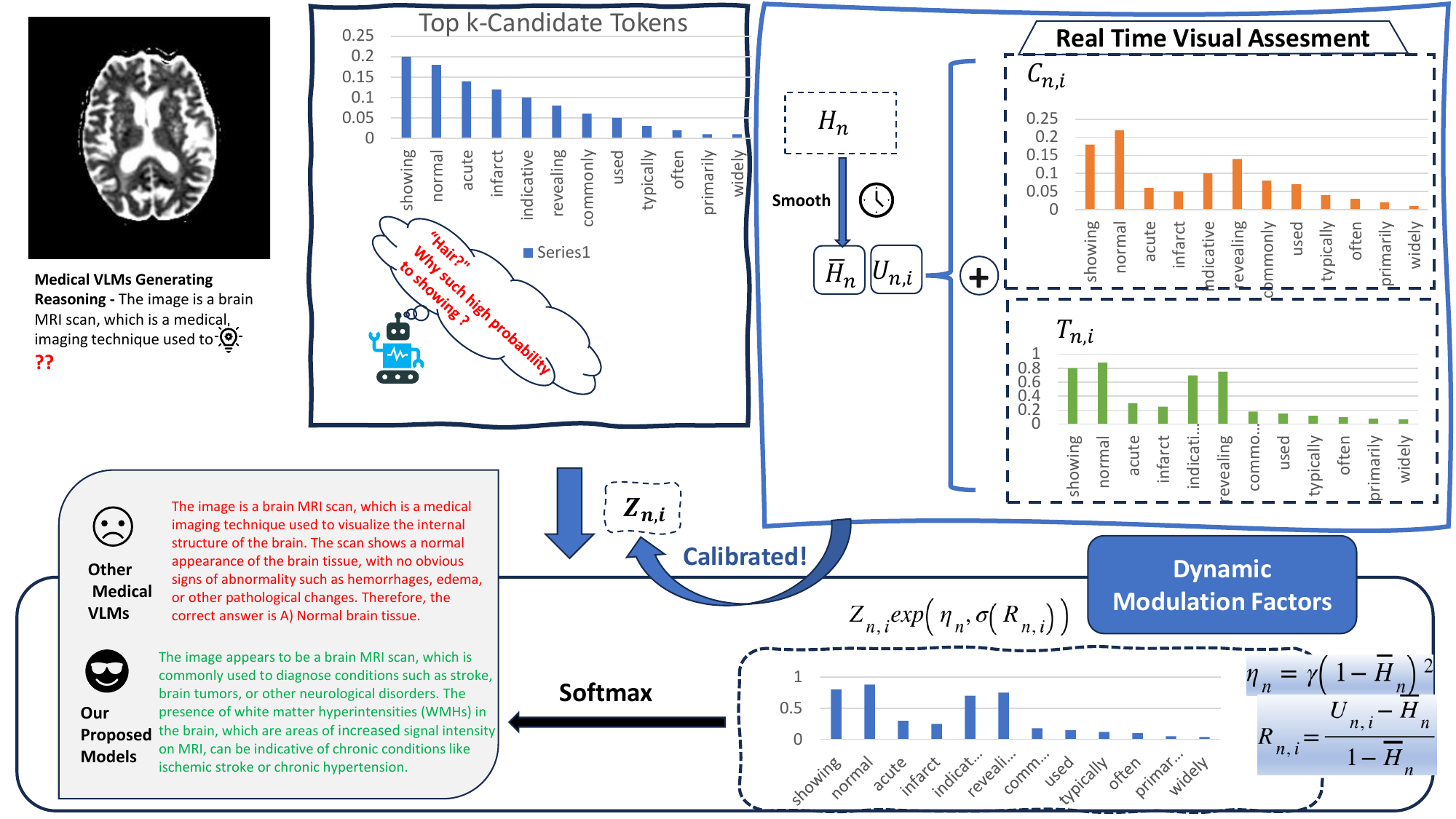}
    \caption{Overview of Proposed I$^2$-Med Model.}
    \label{Fig:2}
\end{figure}

We propose \textbf{I$^2$-Med}, a unified interpretable medical inference framework that enforces structured reasoning and visual faithfulness in medical VLMs. Unlike methods restricted to reinforcement learning or decoding-time correction, it employs a dual-stage design: (i) reward-driven structured reasoning during post-training and (ii) Visual-Guided Logits Regulation (VGLR) at inference, as demonstrated in Figure \ref{Fig:2}. To ensure clinical feasibility, we keep the pretrained medical VLM backbone fixed and avoid parameter updates at inference. Unlike existing approaches that require expensive retraining or fine-tuning, our method does not train the underlying VLM. Instead, interpretability is achieved exclusively during inference through policy optimization and decoding-time visual alignment. By leaving the pretrained model completely unchanged, our framework remains model-agnostic, computationally efficient, and readily deployable with any off-the-shelf VLM, enabling interpretable adaptation to medical tasks without additional training.

\subsection{Structured Policy Optimization for Medical Reasoning}
To promote structured, clinically coherent reasoning without costly chain-of-thought annotations, we adopt a group-relative policy optimization strategy inspired by reinforcement learning–based alignment. Unlike supervised fine-tuning, which maximizes target likelihood, this method uses reward-driven updates to encourage diverse reasoning paths and structured intermediate explanations.\\


\noindent \textbf{Optimization Objective of GRPO:}
Here, we discuss the optimization problem of GRPO, which we directly use in our proposed method. Let $\mathcal{D}$ denote the training dataset and $d \sim \mathcal{D}$ a sampled medical question. Let $p_{\text{old}}$ and $p_{\text{new}}$ represent the previous and updated policies, and $p_{\text{ref}}$ a fixed reference policy from the pretrained backbone. For each question, we sample $K$ candidate responses ${r_i}{i=1}^{K}$ from $p{_\text{old}}(r \mid d)$.


The structured policy optimization objective is defined as:

\begin{align*}
\begin{aligned}
\mathcal{L}_{\text{SPO}} 
=
\mathbb{E}_{d \sim \mathcal{D},\, r_i \sim p_{\text{old}}}
\Bigg[
\frac{1}{K}
\sum_{i=1}^{K}
\Big(
\min
\big(
\rho_i A_i,
\operatorname{clip}(\rho_i, 1-\tau, 1+\tau) A_i
\big)
-
\lambda\, D_{\mathrm{KL}}(p_{\text{new}} \,\|\, p_{\text{ref}})
\Big)
\Bigg],
\end{aligned}
\end{align*}

where
$\rho_i = \frac{p_{\text{new}}(r_i \mid d)}{p_{\text{old}}(r_i \mid d)}$ denotes the policy ratio, $\tau$ is the clipping threshold, and $A_i$ represents the group-normalized advantage.

Instead of using a learned critic, we compute $A_i$ by normalizing rewards within each sampled response group. This group-relative approach improves training stability and reduces computational cost while enabling effective policy updates. A KL divergence term regularizes learning, limiting deviation from the pretrained reference model and preserving multimodal knowledge.

\textbf{Reward Formulation}
To ensure interpretable and diagnostically meaningful outputs, we employ a two-component hierarchical rule-based reward:

\textbf{(1) Structural Consistency Reward.}
The model must place intermediate reasoning within \texttt{<think>} tags and the final decision within \texttt{<answer>} tags. A reward of 1 is assigned when this format is correctly followed, promoting organized and transparent reasoning.

\textbf{(2) Diagnostic Accuracy Reward.}
A reward of 1 is given when the predicted answer matches the ground-truth multiple-choice label. A response is considered correct if the first token inside the \texttt{<answer>} tag corresponds to the correct option (e.g., “A”, “B”, “C”, or “D”).

The total reward is the sum of structural and accuracy components, ensuring both interpretability and diagnostic correctness.
\subsection{Medical Visual-Guided Logits Regulation}
To mitigate visual hallucination and ensure clinical faithfulness, we introduce \emph{Medical Visual-Guided Logits Regulation (MVLR)}. MVLR employs BioMedCLIP as a domain-specific alignment model to enforce medically grounded consistency between radiological images and generated text. Operating entirely at inference, it requires no additional training or parameter updates and only one forward pass per decoding step, improving clinical consistency without added computational cost.

\noindent \textbf{Contextual Medical Alignment Estimation:} At decoding step $n$, prior to selecting token $w_n$, we evaluate the visual consistency of candidate tokens relative to the input medical image $x$. This process begins by estimating the visual grounding of the recent textual context.

\noindent \textbf{Contextual Medical Alignment (CMA):}
We measure the alignment between the input image $x$ and a fixed window of the most recent $M$ generated tokens $w_{n-M:n-1}$ using BioMedCLIP: $S^{\text{CMA}}_{n} = \mathrm{BioMedCLIP}(x, w_{n-M:n-1})$. This score quantifies how well the current reasoning context remains grounded in the visual evidence. To stabilize short-term fluctuations, we maintain a smoothed historical baseline $\bar{S}_{n} = \frac{1}{M} \sum_{j=n-M}^{n-1} S^{\text{CMA}}_{j}$, which represents the established level of visual grounding in recent generation steps.

\noindent \textbf{Candidate-Level Visual Assessment:} For the top-$k$ candidate tokens $\{u_1, \dots, u_k\}$ proposed by the medical VLM at step $n$, we compute two complementary alignment measures.
\begin{enumerate}
    \item \textbf{Conditional Clinical Consistency (CCC):}
This measures visual alignment when appending a candidate token to the current context: $S^{\text{CCC}}_{n,i} = \mathrm{BioMedCLIP}(x, w_{n-M:n-1} \oplus u_i)$, where $\oplus$ denotes concatenation.
\item \textbf{Token Visual Affinity (TVA):}
This measures the intrinsic alignment between the image and the candidate token: $S^{\text{TVA}}_{n,i} = \mathrm{BioMedCLIP}(x, u_i)$.
\end{enumerate}

Since contextual and intrinsic alignments capture complementary aspects of visual grounding, we compute a unified medical grounding score:
\begin{equation}
G_{n,i} = \frac{S^{\text{CCC}}_{n,i} + S^{\text{TVA}}_{n,i}}{2}.
\end{equation}
\noindent This score provides a robust estimate of each candidate’s clinical consistency.\\

\noindent \textbf{Adaptive Clinical Logit Regulation} Having computed the contextual baseline $\bar{S}_n$ and candidate grounding scores $G_{n,i}$, we adaptively regulate the original logits $Z_{n,i}$ produced by the VLM to favor clinically grounded tokens.

\noindent \textbf{Relative Clinical Grounding (RCG)}
We quantify how much each candidate improves or degrades visual alignment relative to the recent baseline: $R_{n,i} = \frac{G_{n,i} - \bar{S}_n}{1 - \bar{S}_n}$. A positive value indicates stronger grounding than the recent contextual baseline.

\noindent \textbf{Adaptive Clinical Constraint (ACC).}
The strength of intervention should depend on the current state of grounding. When the model is already well aligned with the image, aggressive regulation is unnecessary. Conversely, stronger correction is required when grounding weakens. We therefore define: $\alpha_n = \gamma (1 - \bar{S}_n)^2$, where $\gamma$ controls the maximum modulation intensity.\\

\noindent \textbf{Logit Update Rule.}
Finally, the calibrated logit is computed as: $Z'_{n,i} = Z_{n,i} \cdot \exp\!\left(\alpha_n \cdot \sigma(R_{n,i})\right)$, where $\sigma(\cdot)$ denotes the sigmoid function.

This multiplicative regulation increases the probability of clinically consistent tokens while suppressing visually unsupported ones. By incorporating domain-specific visual grounding into decoding, MVLR reduces semantic drift and hallucination without affecting fluency or inference efficiency.

\begin{table*}[t]


\centering
\setlength{\tabcolsep}{10pt}

\caption{\small{Comparison of performance across eight medical modalities against baseline models.}}
\label{Table: 2}
\resizebox{0.8\textwidth}{!}{%
\begin{tabular}{l|cccccccc|c}
\hline
\diagbox{Methods}{Modality}
& CT & MRI & X-Ray & US & DER & FP & OCT & MICRO & Overall \\
\hline
\multicolumn{10}{c}{\textbf{Zero-shot VLMs}} \\
\hline
BLIP-2$^{\dagger}$~\cite{pmlr-v202-li23q}           & 56.74 & 41.32 & 67.58 & 37.27 & 40.65 & 46.24 & 68.08 & 50.40 & 51.04 \\
InstructBLIP$^{\dagger}$~\cite{NEURIPS2023_9a6a435e}     & 28.72 & 33.15 & 61.04 & 41.25 & 62.22 & 50.31 & 42.59 & 46.29 & 45.70 \\
LLaVA$^{\dagger}$~\cite{NEURIPS2023_6dcf277e}         & 17.73 & 26.72 & 30.70 & 18.66 & 49.74 & 47.11 & 33.73 & 28.87 & 31.66 \\
LLaMA Adapter v2$^{\dagger}$~\cite{gao2023llamaadapterv2parameterefficientvisual} & 21.41 & 26.63 & 46.44 & 34.05 & 51.76 & 50.74 & 33.00 & 38.66 & 37.83 \\
MiniGPT-4$^{\dagger}$ [30]        & 22.81 & 27.48 & 38.30 & 25.50 & 40.25 & 38.33 & 31.40 & 36.23 & 32.54 \\
InternVL2 [31]                    & 40.20 & 58.10 & 57.90 & 49.10 & 51.90 & 53.20 & 59.10 & 64.00 & 54.19 \\
Qwen2-VL-2B [3]                   & 45.10 & 38.57 & 39.32 & 30.86 & 35.83 & 43.17 & 35.14 & 36.85 & 38.11 \\
Qwen2-VL-7B [3]                   & 61.46 & 45.77 & 64.27 & 36.01 & 49.08 & 59.84 & 59.32 & 61.08 & 54.60 \\
Qwen2-VL-72B [32]                 & 67.97 & 69.39 & 77.21 & 51.39 & 65.31 & 72.58 & 72.76 & 67.83 & 68.05 \\
Qwen2.5-VL-3B [33]                & 53.87 & 54.23 & 61.84 & 32.69 & 52.94 & 62.47 & 56.23 & 59.64 & 54.24 \\
Qwen2.5-VL-7B [33]                & 60.44 & 58.44 & 73.99 & 30.66 & 62.48 & 67.30 & 61.20 & 67.84 & 60.29 \\
Qwen2.5-VL-72B [33]               & 66.18 & 68.74 & 77.59 & 49.81 & 69.75 & 71.04 & 69.22 & 69.37 & 67.71 \\
\hline
\multicolumn{10}{c}{\textbf{Zero-shot Medical VLMs}} \\
\hline
LLaVA-Med$^{\dagger}$ [8]         & 18.69 & 27.47 & 30.68 & 29.88 & 44.95 & 39.03 & 34.61 & 33.29 & 32.33 \\
RadFM$^{\dagger}$ [34]            & 27.56 & 24.06 & 30.95 & 16.57 & 39.21 & 36.89 & 32.80 & 27.97 & 29.50 \\
Med-Flamingo$^{\dagger}$ [6]      & 38.47 & 40.56 & 30.34 & 24.64 & 32.43 & 30.12 & 26.51 & 19.93 & 30.38 \\
MedVInT$^{\dagger}$ [7]           & 40.74 & 43.10 & 55.10 & 41.26 & 29.11 & 31.84 & 23.26 & 32.00 & 37.05 \\
HuatuoGPT-Vision [9]              & 35.30 & 40.40 & 41.50 & 60.10 & 53.10 & 51.40 & 59.30 & 62.30 & 50.43 \\
HealthGPT [35]                    & 35.50 & 78.50 & 81.90 & 51.40 & 64.90 & 54.60 & 89.30 & 88.20 & 68.04 \\
\hline
\multicolumn{10}{c}{\textbf{Fine-tuned VLMs}} \\
\hline
Qwen2-VL-2B (SFT)                 & 51.74 & 52.83 & 65.57 & 47.65 & 51.91 & 52.26 & 53.99 & 56.58 & 54.07 \\
Qwen2.5-VL-3B (SFT)               & 56.06 & 60.81 & 69.23 & 41.77 & 60.11 & 69.19 & 63.95 & 65.66 & 60.85 \\
Qwen2-VL-2B (Think)               & 66.30 & 71.67 & 77.73 & 57.31 & 72.33 & 71.20 & 71.96 & 70.80 & 69.91 \\
Qwen2-VL-2B (Think) (256 Token)              & 58.16 & 64.98 & 70.48 & 50.68 & 63.69 & 66.2 & 60.96 & 65.64 & 62.60 \\
\hline
Proposed & 60.21  & 65.29   & 71.10  & 51.75  & 66.42  & 69.32  & 63.79  & 66.41  & 65.29  \\
\hline
\end{tabular}%
}

\end{table*}

\begin{table*}[t]





\centering
\setlength{\tabcolsep}{10pt}

\caption{Comparison of our proposed models with medical VLMs baselines on five medical VQA tasks across five clinical reasoning types, evaluated under general-purpose zero-shot, medical zero-shot, and fine-tuned models.}

\label{Table: 4}
\resizebox{\columnwidth}{!}{%
\begin{tabular}{l|ccccc|c}
\hline
\diagbox{Methods}{Types}
& Anatomy Identification
& Disease Diagnosis
& Lesion Grading
& Modality Recognition
& Other Attributes
& Overall \\
\hline

\multicolumn{7}{c}{\textbf{Zero-shot VLMs}}\\
\hline
BLIP-2 & 44.39 & 44.51 & 29.03 & 68.19 & 67.95 & 48.12 \\
InstructBLIP & 44.35 & 32.29 & 59.25 & 75.27 & 23.72 & 40.40 \\
LLaVA & 25.86 & 29.10 & 43.95 & 21.36 & 31.90 & 27.96 \\
LLaMA Adapter v2 & 33.72 & 31.19 & 41.99 & 37.29 & 34.22 & 32.82 \\
MiniGPT-4 & 28.88 & 30.47 & 34.56 & 26.43 & 30.36 & 29.74 \\
Qwen2-VL-2B & 30.70 & 36.53 & 43.58 & 59.90 & 42.19 & 42.58 \\
Qwen2-VL-7B & 42.57 & 48.83 & 52.06 & 84.74 & 59.66 & 57.57 \\
Qwen2-VL-72B & 56.41 & 65.71 & 62.15 & 98.11 & 80.53 & 72.58 \\
Qwen2.5-VL-3B & 35.23 & 50.45 & 52.79 & 85.23 & 54.77 & 55.69 \\
Qwen2.5-VL-7B & 41.00 & 61.32 & 54.13 & 97.78 & 70.45 & 64.94 \\
Qwen2.5-VL-72B & 57.22 & 62.55 & 60.32 & 98.20 & 77.41 & 71.14 \\
\hline
\multicolumn{7}{c}{\textbf{Zero-shot Medical VLMs}}\\
\hline
LLaVA-Med & 29.53 & 29.22 & 34.18 & 26.93 & 33.08 & 29.25 \\
RadFM & 13.31 & 21.69 & 30.35 & 26.64 & 43.85 & 26.99 \\
Med-Flamingo & 24.93 & 38.90 & 30.74 & 30.19 & 14.18 & 34.03 \\
MedVInT & 40.26 & 35.78 & 12.77 & 68.10 & 30.30 & 40.04 \\
\hline
\multicolumn{7}{c}{\textbf{Fine-tuned VLMs}}\\
\hline
Qwen2-VL-2B (SFT) & 53.97 & 51.62 & 60.71 & 86.77 & 63.91 & 63.39 \\
Qwen2.5-VL-3B (SFT) & 54.91 & 57.75 & 64.04 & 84.95 & 71.56 & 66.64 \\
Qwen2-VL-2B (Think) & 62.88 & 66.08 & 65.87 & 98.24 & 80.14 & 74.64 \\
Qwen2-VL-2B (Think) (256 Token)              & 55.35 & 62.61 & 66.47 & 88.29 & 72.78 & 69.10  \\
\hline
Proposed &  56.00 & 64.03 & 66.43 & 88.96 & 73.58 & 69.80\\
\hline
\end{tabular}%
}


\end{table*}

\section{Experiments}
This section presents the datasets and results. We use the open-access dataset OmniMedVQA benchmark~\cite{hu2024omnimedvqa}, a large-scale medical VQA dataset, for evaluation. Further implementation details are provided on GitHub.

\subsection{Results Across Eight Medical Imaging Modalities}
We evaluate I$^2$-Med across eight imaging modalities: CT, MRI, X-ray, Ultrasound, Dermoscopy, Fundus Photography, OCT, and Microscopy. Results on these eight modalities, with baseline comparisons are reported in Table~\ref{Table: 2}. For controlled and efficient evaluation, reasoning length is limited to 256 tokens. We compare against three baseline groups: general-purpose zero-shot VLMs, zero-shot medical VLMs, and fine-tuned VLMs. I$^2$-Med consistently outperforms both zero-shot categories and achieves competitive, often superior, performance relative to fine-tuned models. Although Qwen2-VL-2B (Think) achieves slightly higher accuracy with unrestricted reasoning, it relies on substantially longer outputs and greater computational cost. Under the 256-token constraint, I$^2$-Med surpasses Qwen2-VL-2B (Think) (256 Token) in overall accuracy while maintaining structured, clinically grounded reasoning. Some cross-modality results have been discussed on our GitHub. These results demonstrate strong cross-modality generalization with improved efficiency, indicating that structured reasoning with medical visual grounding is more effective than merely increasing generation length.

\subsection{Results Across Five Clinical Task Categories}

We evaluate I$^2$-Med on five clinical tasks: Anatomy Identification, Disease Diagnosis, Lesion Grading, Modality Recognition, and Other Biological Attributes. The results are presented in Table~\ref{Table: 4}. It achieves stable, competitive results, surpassing Qwen2-VL-2B (Think) (256 Token) in overall average under the 256-token limit. The slight 0.04 difference in Lesion Grading does not affect overall consistency. Detailed cross-modality results for these five clinical task categories are provided in our GitHub.


\begin{figure}[t]
    \centering
\includegraphics[width=1\linewidth]{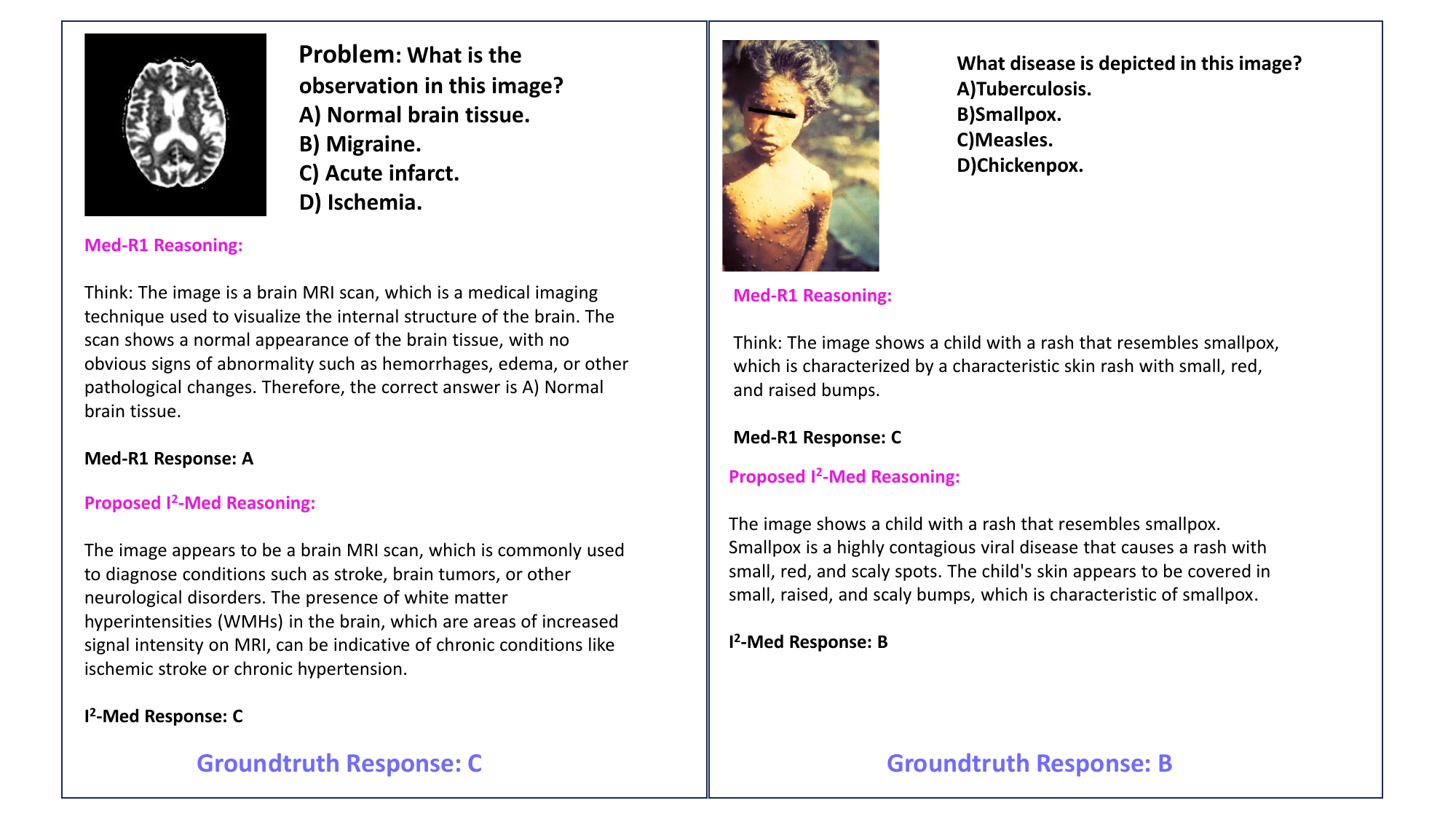}
    \caption{Comparison of Med-R1 and I$^2$-Med on medical VQA examples shows that while both generate explicit reasoning, I$^2$-Med provides more clinically grounded explanations and achieves more accurate predictions.}
    \label{Interpretability}
\end{figure}

\subsection{Interpretability: Qualitative Analysis of Generated Reasoning on Medical VQA}
In Figure~\ref{Interpretability}, we qualitatively compare $I^2$-Med with the Med-R1 baseline. In the first example (left), Med-R1 correctly identifies the modality but misses subtle pathological cues, leading to a false negative. In contrast, $I^2$-Med detects white matter hyperintensities and localizes the acute infarct, demonstrating more accurate and clinically grounded reasoning. In the second example (right), Med-R1 recognizes smallpox-related features but incorrectly predicts “Measles”. $I^2$-Med maintains logical consistency, correctly associating umbilicated pustules with Smallpox. This shows that $I^2$Med effectively leverages its reasoning steps to reach the correct conclusion, whereas Med-R1 fails to translate its reasoning into an accurate prediction. Additional images related to qualitative analysis and post hoc interpretability are available on GitHub.

\section{Conclusion}
In this work, we proposed I$^2$-Med, an interpretable medical inference framework that enhances visual faithfulness in medical vision-language models through inference-time visual-guided logits calibration. By integrating structured policy optimization with dynamic token-level visual grounding, our method effectively reduces visual hallucination without requiring additional training or parameter updates. Extensive evaluations on the OmniMedVQA benchmark across eight imaging modalities and five clinical tasks demonstrate improved reasoning quality, clinical consistency, and competitive accuracy. Qualitative analyses further confirm that I$^2$-Med generates more diagnostically grounded and coherent explanations. Overall, our framework advances trustworthy and clinically deployable medical VLMs. Future work will investigate the effectiveness of our approach on a wider variety of medical VLM benchmarks and real-world clinical applications.

\vspace{2mm}

\noindent\textbf{Acknowledgments.} The work of Md Sajid is supported by the Council of Scientific and Industrial Research (CSIR), New Delhi, for providing fellowship under the Grant 09/1022(13847)/2022-EMR-I. This work is supported by PraxiaTech Private Limited. Further, this work is supported by IITI DRISHTI CPS Foundation under the National Mission on Interdisciplinary Cyber Physical System (NM-ICPS) of the Department of Science and Technology, Government of India.

\vspace{2mm}

\noindent\textbf{Disclosure of Interests.} The authors have no competing interests.

\bibliographystyle{splncs04}
\bibliography{refs}

\end{document}